\documentclass[10pt]{article} % For LaTeX2e
\usepackage[preprint]{tmlr}

\usepackage{amsmath,amsfonts,bm}

\def\eqref#1{equation~\ref{#1}}
\def\1{\bm{1}}

\DeclareMathAlphabet{\mathsfit}{\encodingdefault}{\sfdefault}{m}{sl}
\SetMathAlphabet{\mathsfit}{bold}{\encodingdefault}{\sfdefault}{bx}{n}

\usepackage{hyperref}
\usepackage{graphicx}
\usepackage{amssymb}
\usepackage{amsmath}
\usepackage{mathtools}
\usepackage{amsthm}
\usepackage{algorithm}
\usepackage{gensymb}
\usepackage{algorithmic}
\theoremstyle{definition}
\newtheorem{theorem}{Theorem}
\newtheorem{proposition}{Proposition}

\newtheorem{remark}{Remark}
\usepackage{chngcntr} 
\usepackage{url}
\usepackage{cleveref} 
\usepackage{makecell} 
\usepackage{subcaption}
\usepackage{wrapfig}

\usepackage{xspace}  % optional but improves spacing
\newcommand{\ie}{i.e.,\@\xspace}
\title{Information-Guided Diffusion Sampling for Dataset Distillation}

\author{
  \name\hspace{-1ex}Linfeng Ye$^{1}$ \quad Shayan Mohajer Hamidi$^{2}$ \quad Guang Li$^{3}$\thanks{Correspondence to: Guang Li <guang@lmd.ist.hokudai.ac.jp>} \\ Takahiro Ogawa$^{3}$ \quad Miki Haseyama$^{3}$ \quad Konstantinos N. Plataniotis$^{1}$ \\
  \addr $^{1}$University of Toronto \quad $^{2}$Stanford University \quad $^{3}$Hokkaido University \\
}

\def\month{MM}  % Insert correct month for camera-ready version
\def\year{YYYY} % Insert correct year for camera-ready version
\def\openreview{\url{https://openreview.net/forum?id=XXXX}} % Insert correct link to OpenReview for camera-ready version

\begin{document}

\maketitle

\begin{abstract}
Dataset distillation aims to create a compact dataset that retains essential information while maintaining model performance. Diffusion models (DMs) have shown promise for this task but struggle in low images-per-class (IPC) settings, where generated samples lack diversity. In this paper, we address this issue from an information-theoretic perspective by identifying two key types of information that a distilled dataset must preserve: ($i$) \textit{prototype information} $\mathrm{I}(X;Y)$, which captures label-relevant features; and ($ii$) \textit{contextual information} $\mathrm{H}(X | Y)$, which preserves intra-class variability. Here, $(X,Y)$ represents the pair of random variables corresponding to the input data and its ground truth label, respectively. Observing that the required contextual information scales with IPC, we propose maximizing $\mathrm{I}(X;Y) + \beta \mathrm{H}(X | Y)$ during the DM sampling process, where $\beta$ is IPC-dependent. Since directly computing $\mathrm{I}(X;Y)$ and $\mathrm{H}(X | Y)$ is intractable, we develop \textit{variational estimations} to tightly lower-bound these quantities via a data-driven approach.  Our approach, information-guided diffusion sampling (IGDS), seamlessly integrates with diffusion models and improves dataset distillation across all IPC settings. Experiments on Tiny ImageNet and ImageNet subsets show that IGDS significantly outperforms existing methods, particularly in low-IPC regimes. The code will be released upon acceptance.
\end{abstract}

\section{Introduction}

\label{sec:intro}

The success of high-performance deep neural networks (DNNs) is largely attributed to large-scale, highly informative datasets \citep{lecun2015deep}. However, the size of these datasets poses a substantial burden on storage and computational resources during model training \citep{5206848,10585292, kaplan2020scaling}.
To mitigate the cost of training DNNs, dataset distillation \citep{wang2018dataset, sachdeva2023data} has been extensively studied in recent years as a potential solution to compress datasets, thereby reducing both storage requirements and computational costs. In this approach, a smaller dataset whose compactness is typically measured by images-per-class (IPC) is constructed as a substitute for the original dataset, while still enabling the trained model to achieve decent generalization performance on unseen test data points.

To construct such a compact dataset, the distillation process typically involves an iterative optimization of pixel values and auxiliary model weights to align either with the model's weight trajectory \citep{zhao2021dataset, cazenavette2022dataset, li2024iadd} or feature statistics \citep{wang2022cafe, deng2024exploiting, sajedi2023datadam, li2025hyperbolic}. However, this approach has two major drawbacks: ($i$) High computational cost---most existing methods require jointly optimizing auxiliary model parameters and distilled samples at the pixel level through an iterative process, resulting in significant computational overhead. ($ii$) Poor generalization across different model architectures---the performance of models trained on the distilled dataset is highly dependent on the architecture of the auxiliary DNN. When the target model's architecture differs from that of the auxiliary DNN, significant performance degradation is often observed.

To overcome these drawbacks, recent studies have proposed generative distillation \citep{zhao2022synthesizing, cazenavette2023generalizing}, which leverages a generative model to synthesize a new, compact dataset. In this approach, a generative model is trained on the target dataset and then used to sample distilled data \citep{zhao2022synthesizing, li2024generative, li2025generative}. Consequently, the resulting distilled dataset is both model-architecture-agnostic and more efficiently generated. Among generative models, diffusion models (DMs) have emerged as a strong choice for dataset distillation, demonstrating state-of-the-art performance in this setting \citep{gu2024efficient, su2024diffusion, chen2025influenceguided, li2025diversity}.
Nonetheless, DM-based dataset distillation suffers from poor performance in low-IPC scenarios, where the number of IPCs is small. In these cases, the accuracy is nearly as low as training on a randomly chosen subset. A primary reason is that, under low-IPC conditions, the model’s generated samples reflect only part of the true data distribution, leading to a distilled dataset with limited diversity and substantial information loss. This shortfall grows more severe as the IPC decreases.

To address this limitation and enable the generation of informative samples, we first seek to quantify the essential information that must be preserved. To this end, we adopt an information-theoretic perspective \citep{6773024, 10.5555/1146355, 10619241}. Specifically, we quantify the relevant information using the Shannon entropy $\mathrm{H}(\cdot)$ \citep{6773024} on the random variable (RV) $X$, which represents the input dataset. We then expand $\mathrm{H}(X)$ as: $\mathrm{H}(X) = \mathrm{I}(X;Y) + \mathrm{H}(X | Y)$,
where $Y$ is an RV denoting the ground-truth (GT) label\footnote{We use “ground truth” and “prototype” interchangeably in this paper.}. Through this decomposition, the total information in $\mathrm{H}(X)$ naturally splits into: ($i$) \textit{prototype information} $\mathrm{I}(X;Y)$, reflecting how much information $X$ provides about its GT label; and ($ii$) \textit{contextual information} $\mathrm{H}(X | Y)$, capturing the remaining information in $X$ once its GT label is given. 

A successful dataset distillation should preserve both the prototype and contextual information of the target dataset. However, we observe that the required amount of contextual information depends on the IPC: a higher (resp. lower) IPC necessitates more (resp. less) contextual information. Building on this insight, we propose to maximize $\mathrm{I}(X;Y) + \beta \mathrm{H}(X | Y)$ during the DM sampling process, where $\beta$ is selected according to the IPC. 
Since directly computing $\mathrm{I}(X;Y)$ and $\mathrm{H}(X | Y)$ is intractable, we develop \textit{variational estimations} to tightly lower-bound these quantities via a data-driven approach. Specifically, we train a DNN using a novel training algorithm that provides tight lower bounds on both $\mathrm{I}(X;Y)$ and $\mathrm{H}(X|Y)$. We refer to this DNN as a variational estimator (VE). Once the VE is trained, it is frozen and used in the DM sampling process, guiding the generation of distilled data that maximally preserves both prototype and contextual information.

Our work introduces a novel dataset distillation approach, Information-Guided Diffusion Sampling (IGDS), which leverages information-theoretic principles to enhance the effectiveness of diffusion models in low IPC settings. The key contributions of this paper are as follows:

\noindent $\bullet$ We introduce a principled framework based on Shannon entropy decomposition, identifying prototype information $\mathrm{I}(X;Y)$ and contextual information $\mathrm{H}(X | Y)$ as crucial components for effective dataset distillation. Our approach dynamically balances these terms to optimize the informativeness of the distilled dataset.

\noindent $\bullet$ Since directly computing prototype information $\mathrm{I}(X;Y)$ and contextual information $\mathrm{H}(X | Y)$ is intractable, we develop a data-driven VE using deep neural networks to obtain tight lower bounds on these quantities. This estimator is seamlessly integrated into the diffusion sampling process.

\noindent $\bullet$ We propose IGDS, a novel diffusion-based dataset distillation method that maximizes $\mathrm{I}(X;Y) + \beta \mathrm{H}(X | Y)$ during the sampling process. The weight $\beta$ is IPC-dependent, allowing for adaptive control over prototype and contextual information retention.

\noindent $\bullet$ Extensive experiments on Tiny ImageNet \citep{le2015tiny} and subsets of ImageNet \citep{5206848} demonstrate that IGDS achieves superior performance compared to existing methods, particularly in low-IPC scenarios, where prior diffusion-based approaches suffer from poor diversity and high information loss.

\section{Related Works}
\label{sec:RelatedWorks}
Dataset distillation has received widespread attention since it was proposed, and a substantial amount of research has contributed to its rapid development \citep{li2022awesome, yu2023review}. The proposed methods can be classified into non-generative and generative approaches. 

\subsection{Non-generative Dataset Distillation}
This approach initializes the distilled dataset and optimizes the images during the distillation process using specific algorithms such as matching-based and kernel-based methods \citep{nguyen2021kip}. Matching-based methods ensure the ability to distill by matching parameters, features, or distributions between the original and distilled datasets. For instance, methods like DC \citep{zhao2021datasetcondensation} and IDC \citep{kim2022IDC} match the gradient obtained by training both on original and synthetic data. While approaches like MTT \citep{cazenavette2022MTT} and ATT \citep{liu2024ATT} achieve parameter matching by minimizing the loss over the training trajectory on original and synthetic data. Kernel-based methods take ridge regression as the optimization objective and employ neural tangent kernel to generate distilled datasets \citep{nguyen2021kernel2}. 

\subsection{Generative Dataset Distillation}
This method distills the knowledge of the original dataset into the generative model, which is then used to obtain distilled datasets in the subsequent sampling phase \citep{zhao2022synthesizing, cazenavette2023generalizing}. In each distillation process, traditional methods optimize one dataset of a specific size, as defined by the IPC. Generative methods, however, can generate any number of datasets of any size \citep{gu2024minimax, su2024d4m, su2024diffusion}. This flexibility makes dataset distillation free from the constraints of IPC, thus saving a significant amount of time when distillation needs to be executed more than once, which is common in various downstream tasks of dataset distillation, such as continue learning \citep{yang2023continualLearn}, privacy preservation \citep{li2020soft, li2022compressed, li2023sharing}, and neural architecture search \citep{ding2024hyperSearch}.

Among generative approaches, both our proposed IGDS and the recent influence-guided diffusion (IGD) method \citep{chen2025influenceguided} utilize diffusion models, but differ substantially in motivation and methodology. IGD is training-free and guides sampling based on influence functions that quantify downstream model impact. In contrast, IGDS adopts an information-theoretic perspective, maximizing mutual and conditional entropy through variational estimators. This enables IGDS to better preserve diversity under low-IPC constraints, whereas IGD is more effective in high-IPC regimes.

\subsection{Diffusion Model}
Having a certain degree of creativity, generative models have experienced rapid advancement in recent years. Generative approaches like GANs and VAEs have achieved broad application in various industries. Among them, diffusion models like Imagen \citep{saharia2023imagen} and Stable Diffusion \citep{rombach2022high} have gained significant attention. They acquire the ability to recover an image from random noise by learning how to predict noise from noisy images, which understates their recognized performance stability and distillation versatility. Specific to the field of computer vision, they have been proven effective in a wide range of scenarios \citep{yang2023DiffusionSurvey}. For instance, LDM \citep{rombach2022high} enables the diffusion process in the latent space to save computational resources. DDeP \citep{brempong2022DDeP} utilizes text-to-image diffusion models to obtain promising results on semantic segmentation. Palette \citep{saharia2023Palette} tackles several image generation tasks with conditional diffusion models, and FDM \citep{harvey2022FDM} allows for sampling specific video frames from other video subsets. Our method aims to explore the potential of the diffusion model in dataset distillation.

\section{Notation}
For a positive integer $C$,  let $[C]\triangleq \{1,\dots,C\}$. Denote by $P[i]$ the $i$-th element of the vector $P$.  
For two vectors $U$ and $ V$, denote by $\langle U, V\rangle$ their inner product. For two matrices $M\in \mathbb{R}^{m\times n}$ and $N\in \mathbb{R}^{n\times k}$, denote by $M@N$ their matrix product. We use $|\mathcal{C}|$ to denote the cardinality of a set $\mathcal{C}$. 
The entropy of $C$-dimensional probability vector $P$ is defined as $\mathrm{H} (P) = \sum_{c=1}^C -P [c] \log P [c]$. Also, the cross entropy of two $C$-dimensional probability vectors $P_1$ and $P_2$ is defined as $\mathrm{H} (P_1, P_2) = \sum_{c=1}^C -P_1 [c] \log P_2 [c]$, and their Kullback–Leibler ($\mathrm{KL}$) divergence is defined as $\mathrm{KL} (P_1 || P_2 ) = \sum_{c=1}^C P_1 [c] \log {P_1 [c] \over P_2 [c]}$.
\begin{figure}[!t]
  \centering
   \includegraphics[width=0.75\textwidth]{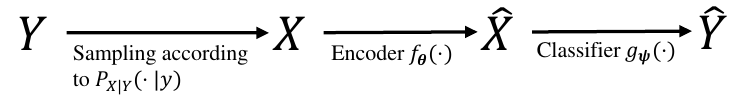}
   \caption{Multi-class classification can be modeled as a Markov chain. Sample $X$ is sampled from the class $Y$, according to the $P_{X|Y}(\cdot|y)$. The encoder then maps the $X$ to the feature representation $\hat{X}$, which is further mapped by the classifier to an output probability vector $\hat{Y}$.}
   \label{fig:MarkovChain}
\end{figure}
For a random variable $X$, denote by $P_{X}$ its probability distribution, and by $E_{X}[\cdot]$ the expected value w.r.t. $X$. For two random variables $X$ and $Y$, denote by $\mathbb{P}_{(X,Y)}$ their joint distribution. The mutual information between two random variables $X$ and $Y$ is written as $\mathrm{I}(X;Y)$, and the conditional mutual information of $X$ and $Y$ given a third random variable $Z$ is $\mathrm{I}(X;Y|Z)$. Consider a dataset $\mathcal{D}$ of size $n$ with $C$ classes, $\mathcal{D} = \{(\boldsymbol{x}_i, y_i)\}_{i=1}^n$, where each $\boldsymbol{x}_i \in \mathbb{R}^d$ and $y_i \in [C]$. For any class $y$, we define $\mathcal{D}_y = \{(\boldsymbol{x}_j,y_j) \in \mathcal{D} | y_j = y\}$ the subset of $\mathcal{D}$ containing all samples with label $y$. Lastly, the softmax operation is denoted by $\sigma(\cdot)$. 

\section{Methodology} \label{sec:method}
% \subsection{Motivation}
As discussed in \cref{sec:intro}, the performance of DM-based dataset distillation degrades significantly when the IPC is small. In such low-IPC conditions, the DM tends to produce samples that represent only a portion of the true data distribution, omitting many essential modes. As a result, the distilled dataset exhibits limited diversity and loses substantial information about the underlying classes. Empirically, this often manifests in accuracies that are comparable to training on a mere random subset of the original dataset. The challenge intensifies with decreasing IPC, since fewer examples per class mean the generative process has even less guidance for reproducing the full range of relevant features. In practice, these shortcomings severely limit the applicability of DM-based distillation for real-world tasks where one cannot afford to collect a large number of samples for each class.

To tackle this limitation, it is crucial to identify the core information that must be retained in the distilled dataset. To this end, we adopt an information-theoretic perspective to rigorously quantify and preserve this information. Specifically, we measure the total information in the input dataset, represented as a random variable $X$, using the Shannon entropy $\mathrm{H}(X)$. Noting that a significant part of the value of a dataset lies in its ability to determine labels, we decompose $\mathrm{H}(X)$ as follows
\begin{align}
\mathrm{H}(X) \;=\; \underbrace{\mathrm{I}(X;Y)}_{\text{prototype information}} 
\;+\; 
 \,\underbrace{\mathrm{H}(X| Y)}_{\text{contextual information}},  
\end{align}
where $Y$ is a random variable denoting the GT label. This decomposition distinctly separates prototype information $\mathrm{I}(X;Y)$, which quantifies how much $X$ reveals about its label, from contextual information $\mathrm{H}(X | Y)$, which captures the variability and richness of the data given its label. In other words, prototype information ensures that the distilled dataset remains discriminative for classification tasks, whereas contextual information guards against the collapse into a narrow subset of features, thus maintaining diversity and nuance (in \cref{sec:semantic}, we visualize the semantic meaning of prototype and contextual information). By explicitly accounting for both these components, we can better capture the data’s essential characteristics, even in low-IPC regimes.
A successful dataset distillation scheme must retain both prototype information, ensuring that each class is accurately characterized, and contextual information, preserving the variety and richness of the underlying data distribution. However, our observations indicate that the requisite amount of contextual information scales with the IPC: higher IPC scenarios allow, and indeed necessitate, more contextual detail, whereas lower IPC settings benefit more from a tighter focus on prototype information (please see \cref{sec:beta} for additional details).

Guided by this insight, we aim to balance these two information types by maximizing the objective
\begin{align} \label{eq:ultimatediff}
 \mathrm{I}(X;Y) \;+\; \beta \, \mathrm{H}(X | Y),    
\end{align}
where the scalar $\beta>0$ is chosen to reflect the IPC: a larger $\beta$ for high-IPC settings increases the emphasis on contextual richness, while a smaller $\beta$ in low-IPC scenarios prioritizes critical prototype information. 

Computing $\mathrm{I}(X;Y)$ and $\mathrm{H}(X|Y)$ is challenging, and to the best of our knowledge, no previous work has accomplished this. To overcome this difficulty, we introduce a novel method in \cref{sec:compute_var} that provides variational estimates for these quantities. Subsequently, in \cref{sec:diff}, we leverage these estimates to guide the sampling process of diffusion models.

\begin{wrapfigure}{r}{0.6\textwidth}
  \begin{minipage}{0.6\textwidth}
  \vspace{-1.0cm}
    \begin{algorithm}[H]
       \caption{Pseudo-code for Training the $f_{\boldsymbol{\theta}}(\cdot)$}
       \label{alg:VE}
    \begin{algorithmic}[1]
       \STATE   {\bfseries Input:} $f_{\boldsymbol{\theta}}, f_{\boldsymbol{m}}$: initialized encoder and momentum encoder, $queue$: dictionary as a queue of $K$ keys, $m$: momentum, $aug$: random augmentation method, $\tau$: temperature and $\lambda>0$.
       \STATE $f_{\boldsymbol{\theta}}$.params = $f_{\boldsymbol{m}}$.params
       \FOR{$x \in D$}
       \STATE  $x_q, x_k$ = aug($x$), aug($x$) 
       \STATE $ q,k = f_{\boldsymbol{\theta}}(x_q), f_{\boldsymbol{m}}(x_k)$.detach() 
       \STATE $H_q, H_k$ = softmax($q$), softmax($k$)
       \STATE $Q = (H_q+ H_k) / 2$
       \STATE $l_{pos}, l_{neg} = \langle q,k \rangle,  q@k^T$ \\
       \STATE logits = cat([$l_{pos}, l_{neg}$], dim=1) \\
       \STATE labels = zeros(N)\\
       \STATE loss = CE (logits / $\tau$, labels) - $\lambda \mathrm{KL}(H_q||Q^Y)$
       \STATE loss.backward()
       \STATE update($f_{\boldsymbol{\theta}}$.params)
       \STATE \small $f_{\boldsymbol{m}}$.params = m × $f_{\boldsymbol{m}}$.params + (1-m) × $f_{\boldsymbol{\theta}}$.params\\
       \STATE \normalsize enqueue(queue, k)
       \STATE dequeue(queue)
       \normalsize \color{black} \ENDFOR
       \STATE  {\bfseries Output:} $f_{\boldsymbol{\theta}}$
    \end{algorithmic}
    \end{algorithm}
    \vspace{-1.0cm}
  \end{minipage}
\end{wrapfigure}

\textit{The proofs for all propositions presented in this paper are deferred to \cref{sec:proofs}.}

\subsection{Variational Estimates for \texorpdfstring{$\mathrm{I}(X;Y)$}{} and \texorpdfstring{$\mathrm{H}(X | Y)$}{}} \label{sec:compute_var}

We employ an auxiliary DNN composed of an encoder $f_{\boldsymbol{\theta}}(\cdot)$ and a classifier $g_{\boldsymbol{\psi}}(\cdot)$ to help us finding variational estimates for both $\mathrm{I}(X;Y)$ and $\mathrm{H}(X|Y)$. The encoder transforms the input $X$ into a feature representation $\hat{X}$, and the classifier maps $\hat{X}$ to a probability vector $\hat{Y}$. In this setup, the random variables $\{Y, X, \hat{X}, \hat{Y}\}$ form a Markov chain in the order shown in \cref{fig:MarkovChain} (see \citet{10900607} for more details). Now, in \cref{sec:compute_I} and \cref{sec:compute_H}, we show how this auxiliary DNN can be leveraged to derive variational estimates for $\mathrm{I}(X;Y)$ and $\mathrm{H}(X |Y)$, respectively. Then, in \cref{Sec:trainVE}, we train both  $f_{\boldsymbol{\theta}}(\cdot)$ and $g_{\boldsymbol{\psi}}(\cdot)$ to give us the estimates. 

\subsubsection{Variational Estimates for \texorpdfstring{$\mathrm{I}(X;Y)$}{}} \label{sec:compute_I}

Equipped with $f_{\boldsymbol{\theta}}(\cdot)$ and classifier $g_{\boldsymbol{\psi}}(\cdot)$, in this section we propose a method to find a variational estimation for $\mathrm{I}(X;Y)$. We start by decomposing $\mathrm{I}(X;Y)$ as follows:
\begin{align} \label{Eq:Decompose_I}
\mathrm{I}(X;Y) = \mathrm{I}(\hat{X}; Y) + \mathrm{I}(Y;X|\hat{X}).
\end{align}

The first term on the right-hand side of \cref{Eq:Decompose_I}, namely $\mathrm{I}(\hat{X}; Y)$, is difficult to compute directly. To overcome this difficulty, we introduce the following Proposition:

\begin{proposition}\label{Lemma:ProofInjectiveEq}
Consider a linear classifier $g_{\boldsymbol{\psi}}: \{\hat{X}\rightarrow \hat{Y}| \hat{Y} = \boldsymbol{\psi} \hat{X}\}$, parameterized by $\boldsymbol{\psi}\in \mathbb{R}^{n\times m}$ with $m\geq n$. If $\boldsymbol{\psi}$ has full column rank, then
\begin{align}
    \mathrm{I}(\hat{X}; Y) \;=\; \mathrm{I}(\hat{Y}; Y).
\end{align}    
\end{proposition}
Now, we can write
\begin{align}
    % \mathrm{I}(X;Y) = \mathrm{I}(\hat{X}; Y) + \mathrm{I}(Y;X|\hat{X})\\
    \mathrm{I}(\hat{Y}; Y) &= \mathrm{H}(Y) - \mathrm{H}(Y|
    \hat{Y})\\ \label{eq:yhat}
    &\geq \mathrm{H}(Y) - \mathrm{H}(Y|
    \hat{Y}, Y) \\ \label{eq:yhat2}
    & = \mathrm{H}(Y) + \mathbb{E}_Y \log P_{Y|
    \hat{Y}},
\end{align}
where the inequality in \cref{eq:yhat} becomes equality if $\mathrm{P}_{Y|\hat{Y}} = \mathrm{P}_{Y}$, i.e., the classifier is Bayes-optimal. Now, the quantities in \cref{eq:yhat2} can be easily computed; specifically, $\mathrm{H}(Y)$ is simply the entropy of the ground truth distribution, which is constant for a given dataset., and $\mathbb{E}_Y \log P_{Y|\hat{Y}}$ is the average of cross-entropy of the output, In practice, we use one-hot probabilities to estimate the Bayes probabilities \citep{ye2024bayes}. 

The second term on the right-hand side of \cref{Eq:Decompose_I}, namely $\mathrm{I}(Y; X | \hat{X})$, is difficult to compute directly. In what follows, we propose to minimize this term so that $\mathrm{I}(\hat{X}; Y)$ forms a tight lower bound on $\mathrm{I}(X;Y)$. To motivate this, we first present \cref{prop:I}.

\begin{wrapfigure}{r}{0.6\textwidth}
  \begin{minipage}{0.6\textwidth}
  \vspace{-1.0cm}
    \begin{algorithm}[H]
       \caption{Pseudo-code of IGDS}
       \label{alg:IGDS}
    \begin{algorithmic}[1]
       \STATE   {\bfseries Input:} The number of iterations $N$, $\boldsymbol{y}$, noise levels $\{ \Tilde{\sigma} \}$, pre-trained encoder $f(\cdot)$ and classifier $g(\cdot)$, $\eta>0$, $\beta>0$, $\tau>0$, IPC n, target label $y$.
       \STATE $\boldsymbol{x}_N \sim \mathcal{N} (\mathbf{0}, \mathbf{I})$
       \FOR{$t=N-1, N-2,\dots,0$}
       \STATE  $\hat{\boldsymbol{s}} \leftarrow \boldsymbol{s}_{\theta} (\boldsymbol{x}_t,t)$ 
       \STATE $ \Tilde{\boldsymbol{x}}_0 \leftarrow \frac{1}{\sqrt{\bar\alpha_t}}\left(\boldsymbol{x}_t+(1-\bar\alpha_t)\hat{\boldsymbol{s}} \right)$
       \STATE $\boldsymbol{z} \sim \mathcal{N} (\mathbf{0}, \mathbf{I})$.
       \STATE  $\boldsymbol{x}_{t-1}^\prime \leftarrow \frac{\sqrt{\alpha_t}(1-\bar{\alpha}_{t-1})}{1 - \bar{\alpha}_t}\boldsymbol{x}_t + \frac{\sqrt{\bar{\alpha}_{t-1}}\beta_t}{1 - \bar{\alpha}_t}\Tilde{\boldsymbol{x}}_0 +  {\tilde{\sigma}_t \boldsymbol{z}}$.
       \STATE $\hat{\boldsymbol{x}}_{t-1} = f(\boldsymbol{x}_{t-1}^\prime)$\\
       \STATE $H_{\hat{\boldsymbol{x}}_{t-1}}=SoftMax({\boldsymbol{x}}_{t-1}/\tau)$ \\
       \STATE $Q_{t-1} = \frac{1}{n}\sum H_{\hat{\boldsymbol{x}}_{t-1}}$
       \STATE $\hat{\boldsymbol{y}}_{t-1}=g({\hat{\boldsymbol{x}}}_{t-1})$ \\
       \STATE $\mathcal{L}_{IGDS} = \mathbb{E}\log P_{\hat{y}|y} + \mathrm{H}(\hat{y}) + \beta \mathrm{KL} (H_{\hat{\boldsymbol{x}}_{t-1}}||Q_{t-1})$\\
       \STATE $\boldsymbol{x}_{t-1} \leftarrow \boldsymbol{x}_{t-1}^\prime + \eta \nabla_{\boldsymbol{x}_{t}}  ~  \mathcal{L}_{IGDS}$. \\
       \normalsize \color{black} \ENDFOR
       \STATE  {\bfseries Output:} $\boldsymbol{x}_0$
    \end{algorithmic}
    \end{algorithm}
    \vspace{-1.5cm}
  \end{minipage}
\end{wrapfigure}

\begin{proposition} \label{prop:I}
     For an encoder $f_{\boldsymbol{\theta}}(\cdot)$ parametrized by $\boldsymbol{\theta}$
     \begin{align}
         \min_{\boldsymbol{\theta}} \mathrm{I}(Y;X|\hat{X}) \equiv \max_{\boldsymbol{\theta}} \mathrm{I}(X;\hat{X}).
     \end{align}
\end{proposition}
As per \cref{prop:I}, we shall train $f_{\boldsymbol{\theta}}(\cdot)$ and  $g_{\boldsymbol{\psi}}(\cdot)$ to maximize $\mathrm{I}(X;\hat{X})$. The details of this training process are provided in \cref{Sec:trainVE}. In this manner, we find a variational estimation for $\mathrm{I}(X;Y)$ which we denote by $\underline{\mathrm{I}}(X;Y)$. Particularly, 
\begin{align} \label{I_VIA}
\underline{\mathrm{I}}(X;Y) = \mathrm{H}(\hat{Y}) + \mathbb{E}_Y \log P_{\hat{Y}|Y}.
\end{align}
\subsubsection{Variational Estimates for \texorpdfstring{$\mathrm{H}(X | Y)$}{}} \label{sec:compute_H}

The term $\mathrm{H}(X|Y)$ can be expanded as
\begin{align}
     \mathrm{H}(X|Y) &= \mathrm{I}(X;\hat{X}|Y) + \mathrm{H}(X|\hat{X},Y). 
\end{align}

To compute $\mathrm{I}(X;\hat{X}|Y)$, we introduce the following Proposition.
\begin{proposition} \label{lem:soft}
Assume that the feature representation $\hat{X}$ has zero mean \citep{7410480, Hinton2015DistillingTK}. Then, $\mathrm{I}(X;\hat{X}|Y) = \mathrm{I}(X;\sigma(\hat{X})|Y)$.    
\end{proposition}
Despite $\mathrm{I}(X;\hat{X}|Y)$, the term $\mathrm{I}(X;\sigma(\hat{X})|Y)$ can indeed be calculated analytically using the same approach as used in \citep{10900607, ye2024bayes}:
\begin{align} \label{Eq:LowerBoundOfCMI}
\mathrm{I}(X;\sigma(\hat{X})|Y)  &= \sum_{y\in[C]} P_Y(y)  ~\mathrm{I}(X;\sigma(\hat{X})|y)\\
 & =  \sum_{y\in[C]} P_Y(y) ~ \mathbb{E}_{X|Y}\mathrm{KL}(\sigma(\hat{X})||Q^y)\\
 & =   \mathbb{E}_{X,Y}\mathrm{KL}(\sigma(\hat{X})||Q^Y),  \label{Eq:CMI}
\end{align}

where $Q^y$ can be computed as $\frac{1}{|D_y|}\sum_{x\in D_y} \sigma(\hat{X})$ \citep{10900607}. 

In addition, the term $\mathrm{H}(X|\hat{X},Y)$ is not easy to compute, so we introduce the following proposition to minimize it such that $\mathrm{I}(X;\hat{X}|Y)$ becomes a tight lower bound for $\mathrm{H}(X|Y)$.

\begin{proposition} \label{prop:H}
     For an encoder $f_{\boldsymbol{\theta}}(\cdot)$ parametrized by $\theta$
     \begin{align}
         \min_\theta \mathrm{H}(X|\hat{X},Y) \equiv \max_\theta \mathrm{I}(X;\hat{X}).
     \end{align}
\end{proposition}
As such, we have found a variational estimation for $\mathrm{H}(X|Y)$ which we denote by $\underline{\mathrm{H}}(X|Y)$. Particularly, 
\begin{align} \label{H_VIA}
\underline{\mathrm{H}}(X|Y) = \mathbb{E}_{X,Y}\mathrm{KL}(\sigma(\hat{X})||Q^Y).
\end{align}

\subsubsection{Training Variational Estimator}
\label{Sec:trainVE}

According to \cref{prop:I} and \cref{prop:H}, obtaining tight lower bounds for $\mathrm{I}(X;Y)$ and $\mathrm{H}(X | Y)$, denoted as $\underline{\mathrm{I}}(X;Y)$ and $\underline{\mathrm{H}}(X | Y)$ respectively, requires training $f_{\boldsymbol{\theta}}(\cdot)$ and $g_{\boldsymbol{\psi}}(\cdot)$ to maximize $\mathrm{I}(X;\hat{X})$. We refer to this DNN, which consists of the concatenation of $f_{\boldsymbol{\theta}}(\cdot)$ and $g_{\boldsymbol{\psi}}(\cdot)$, as the variational estimator (VE) since it provides tight estimations $\underline{\mathrm{I}}(X;Y)$ and $\underline{\mathrm{H}}(X | Y)$. The training procedure for the VE is described in this subsection. To establish the theoretical foundation for this approach, we first introduce the following theorem.

\begin{theorem} \label{th:VE}
Consider the mapping $\hat{X}= f_{\boldsymbol{\theta}}(X)$, where $f_{\boldsymbol{\theta}}(\cdot)$ is an encoder parametrized by $\boldsymbol{\theta}$. Then,  
\begin{align} 
\max_{\boldsymbol{\theta}} \mathrm{I}(X; \hat{X}) \equiv \min_{\boldsymbol{\theta}} \big[\mathrm{H}(\tilde{Y}|\hat{X}) - \mathrm{I}(X;\hat{X}|\tilde{Y})\big],  
\end{align}
where $\tilde{Y}$ is an auxiliary random variable denoting the task type.

\end{theorem}

\begin{wrapfigure}{h}{0.6\textwidth}
    \vspace{-1.5cm}
    \begin{center}
    \includegraphics[width=0.6\textwidth]{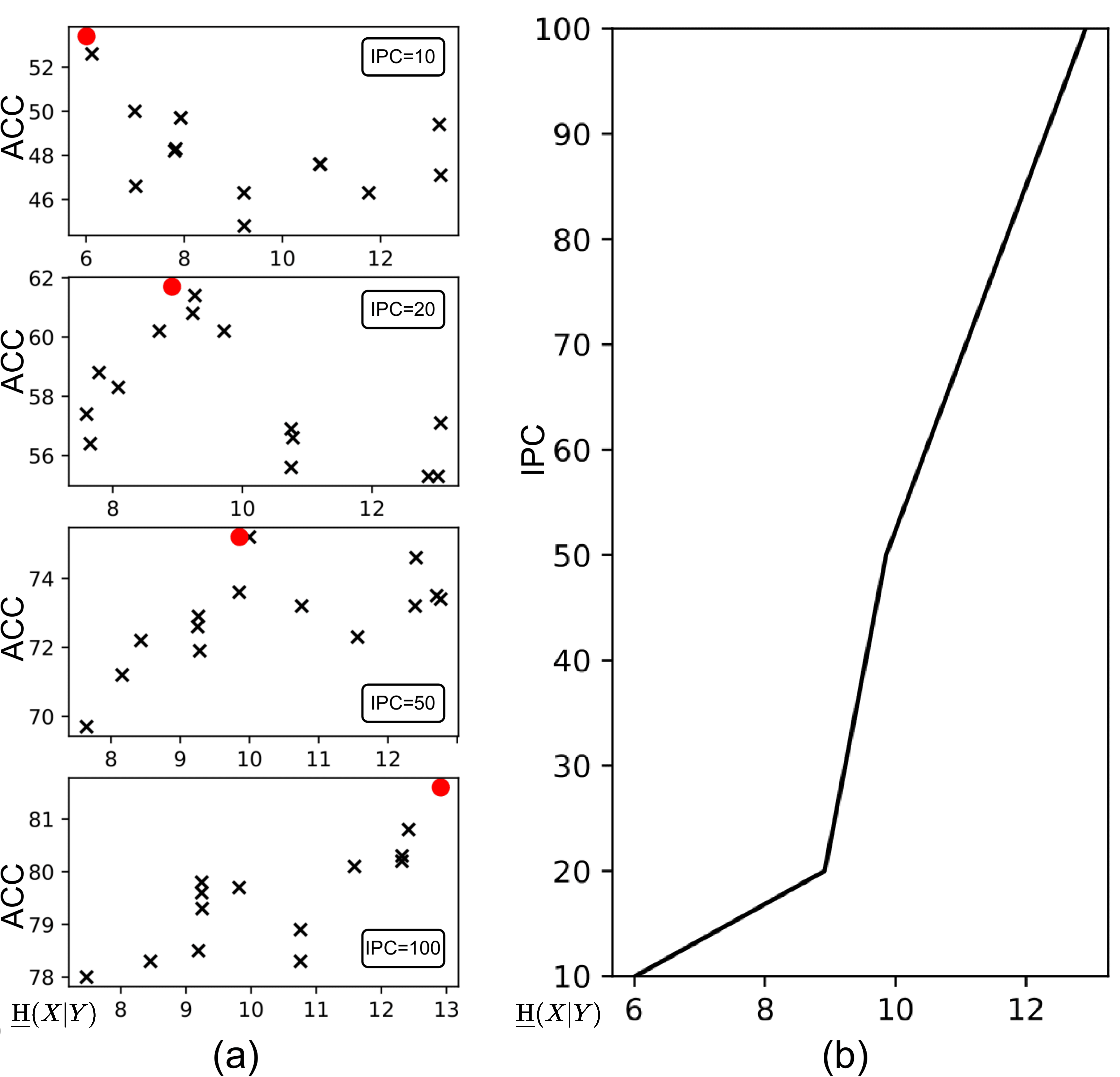}
    \end{center}
     \vspace{-0.5cm}
    \caption{(a) $\underline{\mathrm{H}}(X|Y)$ of selected subsets Vs. model validation accuracy for different IPC settings; the subset associated with the highest model accuracy is marked with a red dot. (b) $\underline{\mathrm{H}}(X|Y)$ of the best subset compared across different IPC settings.}
    \vspace{-1.2cm}
    \label{fig:onecol}
\end{wrapfigure}

    \begin{proof} 
    \text{Using the chain rule, the mutual information } \\
    \text{$\mathrm{I}(X;\tilde{Y}, \hat{X})$ can be expanded in two different ways:}
\begin{subequations} 
\begin{align}
\mathrm{I}(X;\tilde{Y}, \hat{X}) & = \mathrm{I}(X;\tilde{Y}|\hat{X}) + \mathrm{I}(X;\hat{X}) \label{eq:I1}\\
\mathrm{I}(X;\tilde{Y}, \hat{X}) & = \mathrm{I}(X;\hat{X}|\tilde{Y}) + \mathrm{I}(X;\tilde{Y}). \label{eq:I2}
\end{align}    
\end{subequations}
By setting the right hand side of \cref{eq:I1} and \cref{eq:I2} equal to each other, we obtain
\begin{align}
    \mathrm{I}(X; \hat{X}) & = \mathrm{I}(X;\tilde{Y}) -\mathrm{I}(X;\tilde{Y}|\hat{X}) +\mathrm{I}(X;\hat{X}|\tilde{Y}) \\
    & = \mathrm{I}(X;\hat{X}; \tilde{Y}) + \mathrm{I}(X;\hat{X}|\tilde{Y}) \label{Eq:InteractionInfor}\\
    & = \mathrm{I}(\hat{X}; \tilde{Y}) + \mathrm{I}(X;\hat{X}|\tilde{Y}) \label{Eq:InteractionInforMarkov}\\
    & = \mathrm{H}(\tilde{Y}) - \mathrm{H}(\tilde{Y}|\hat{X}) + \mathrm{I}(X;\hat{X}|\tilde{Y}), \label{Eq:decoupleEntropy}
\end{align}
where \cref{Eq:InteractionInfor} follows from the definition of interaction information, and \cref{Eq:InteractionInforMarkov} holds because $\tilde{Y}\rightarrow X\rightarrow \hat{X}$ forms a Markov chain. 
\end{proof}

As per \cref{th:VE}, to maximize $\mathrm{I}(X; \hat{X})$ one can instead maximize the three terms: $\mathrm{H}(\tilde{Y})$,   $-\mathrm{H}(\tilde{Y}|\hat{X})$, and $\mathrm{I}(X;\hat{X}|\tilde{Y})$.
In the following, we discuss how to maximize these three terms.

\noindent $\bullet$ $\mathrm{H}(\tilde{Y})$ depends only on the statistics of the random variable $\tilde{Y}$. In addition, since it is an auxiliary random variable, it can be defined such that $\mathrm{H}(\tilde{Y})$ is maximized. The following Proposition establishes a condition under which $\mathrm{H}(\tilde{Y})$ is maximized.

\begin{proposition} \label{Lemma:MaxHX}
    Given a dataset $\mathcal{D}$, the entropy $\mathrm{H}(\tilde{Y})$ is maximized when each individual sample $x\in \mathcal{D}$ is associated with a unique $\tilde{y}$, and the probability distribution of $\mathrm{P}_{\tilde{Y}} = \frac{1}{|\mathcal{D}|},\ \forall\tilde{y}$.
\end{proposition}
As per \cref{Lemma:MaxHX}, $\mathrm{H}(\tilde{Y})$ is maximized when the task is formulated as an instance discrimination task.

 \noindent $\bullet$ To maximize $-\mathrm{H}(\tilde{Y}|\hat{X})$, we first note that $\mathrm{H}(\tilde{Y}|\hat{X}) = -\sum P(\hat{x},y) \log P(y|\hat{x})$. Since we formulate the problem as an instant discrimination task, and following \citep{galvez2023role, oord2018representation}, we approximate the conditional probability $P(y|\hat{x})$ as
\begin{align}
    P(\tilde{y}_i|\hat{x}_j) \approx \frac{\exp(\langle \hat{x}_i, \hat{x}_j\rangle / \tau)}{\sum_k \exp(\langle \hat{x}_i, \hat{x}_k\rangle / \tau)},
\end{align}
where $\tau$ is a predetermined hyperparameter, usually referred to as the temperature \citep{he2020momentum}. Then, 
\begin{align}
    \mathrm{H}(\tilde{Y}|\hat{X}) &\leq \mathrm{H}(\tilde{Y}|\hat{X}, \hat{Y}|X) =  -\mathbb{E} \log\left[\frac{\exp(\langle \hat{x}_i, \hat{x}_i\rangle / \tau)}{\sum_k \exp(\langle \hat{x}_i, \hat{x}_k\rangle / \tau)}\right], \label{Eq:InfoNCE}
\end{align}

where the expectation is take over $P_{\tilde{Y}|\hat{X}}$. The conditional entropy can be minimized by minimizing \cref{Eq:InfoNCE}.

\noindent $\bullet$ $\mathrm{I}(X;\hat{X}|\tilde{Y})$ can be maximized by iteratively maximizing \cref{Eq:CMI} and updating the $Q^y = \frac{1}{|D_y|}\sum_{x\in D_y} \sigma(\hat{X})$.
\begin{remark}
\label{Remark:MIdecouple1}
  We acknowledge the similarity between VE and information bottleneck theory \citep{tishby2000information, michael2018on}. In \cref{Appendix:VEandBI}, we elaborate on this connection and derive the objective function using an alternative approach.
\end{remark}

\begin{table}[]
\centering
\caption{Comparing model's performance in terms of accuracy on ImageWoof validation set. All results are evaluated under the resolution $256 \times 256$. We use \textbf{bold} number and asterisk ($^*$) to denote the best and the second best results, respectively. }\label{Tab:ImageWoofResults}
\resizebox{1\textwidth}{!}{\begin{tabular}{cc|ccccccccc|c}
\hline
\begin{tabular}
[c]{@{}c@{}}IPC \\ (Ratio)\end{tabular} & Test Model & Random   & \makecell{K-Center }  & \makecell{Herding} & \makecell{DiT} & \makecell{DM}  & \makecell{IDC-1 } & \makecell{GLaD }& \makecell{MiniMax} & \makecell{RDED} & Ours      \\ \hline
$1$                                                      & ConvNet-6      & $14.2_{\pm0.9}$ & $15.6_{\pm1.0}$ & -        & $12.7_{\pm0.6}$ & ${21.1_{\pm0.5}}^*$ & -        & -        & $15.2_{\pm0.6}$ & $18.5 _{\pm 0.9}$ & $\boldsymbol{23.1_{\pm0.8}}$ \\
$(0.08\%)$                                               & ResNetAP-10    & $17.8_{\pm2.4}$ & $18.3_{\pm0.6}$ & -        & $18.0_{\pm1.3}$ & -        & -        & -        & ${18.9_{\pm2.4}}^*$ & -          & $\boldsymbol{23.6_{\pm0.9}}$  \\
                                                       & ResNet-18        & $13.5_{\pm0.4}$ & $12.5_{\pm0.8}$ & -        & $15.3_{\pm0.7}$ & -        & -        & -        & $14.6_{\pm0.6}$ & ${20.8 _{\pm 1.2}}^*$ & $\boldsymbol{22.8_{\pm0.8}}$  \\ \hline
$10$                                                     & ConvNet-6      & $24.3_{\pm1.1}$ & $19.4_{\pm0.9}$ & $26.7_{\pm0.5}$ & $34.2_{\pm1.1}$ & $26.9_{\pm1.2}$ & $33.3_{\pm1.1}$ & $33.8_{\pm0.9}$ & $37.0_{\pm1.0}$ & ${40.6 _{\pm2.0}}^*$  & $\boldsymbol{41.9_{\pm1.5}}$  \\
$(0.4\%)$                                                & ResNetAP-10    & $29.4_{\pm0.8}$ & $22.1_{\pm0.1}$ & $32.0_{\pm0.3}$ & $34.7_{\pm0.5}$ & $30.3_{\pm1.2}$ & $39.1_{\pm0.5}$ & $32.9_{\pm0.9}$ & ${39.2_{\pm1.3}}^*$ & -          & $\boldsymbol{43.5_{\pm0.3}}$  \\
                                                       & ResNet-18        & $27.7_{\pm0.9}$ & $21.1_{\pm0.4}$ & $30.2_{\pm1.2}$ & $34.7_{\pm0.4}$ & $33.4_{\pm0.7}$ & $37.3_{\pm0.2}$ & $31.7_{\pm0.8}$ & $37.6_{\pm0.9}$ & ${38.5 _{\pm 2.1}}^*$ & $\boldsymbol{40.7_{\pm0.5}}$  \\ \hline
$20$                                                     & ConvNet-6      & $29.1_{\pm0.7}$ & $21.5_{\pm0.8}$ & $29.5_{\pm0.3}$ & $36.1_{\pm0.8}$ & $29.9_{\pm1.0}$ & $35.5_{\pm0.8}$ & -        & ${37.6_{\pm0.2}}^*$ & -          & $\boldsymbol{45.7_{\pm0.6}}$  \\
$(1.6\%)$                                                & ResNetAP-10    & $32.7_{\pm0.4}$ & $25.1_{\pm0.7}$ & $34.9_{\pm0.1}$ & $41.1_{\pm0.8}$ & $35.2_{\pm0.6}$ & $43.3_{\pm0.3}$ & -        & ${45.8_{\pm0.5}}^*$ & -          & $\boldsymbol{55.1_{\pm0.6}}$ \\
                                                       & ResNet-18        & $29.7_{\pm0.5}$ & $23.6_{\pm0.3}$ & $32.2_{\pm0.6}$ & $40.5_{\pm0.5}$ & $29.8_{\pm1.7}$ & $38.6_{\pm0.2}$ & -        & ${42.5_{\pm0.6}}^*$ & -          & $\boldsymbol{49.9_{\pm0.7}}$  \\ \hline
$50$                                                     & ConvNet-6      & $41.3_{\pm0.6}$ & $36.5_{\pm1.0}$ & $40.3_{\pm0.7}$ & $46.5_{\pm0.8}$ & $44.4_{\pm1.0}$ & $43.9_{\pm1.2}$ & -        & $53.9_{\pm0.6}$ & ${61.5_{\pm 0.3}}^*$ & $\boldsymbol{65.3_{\pm1.4}}$ \\
$(3.8\%)$                                                & ResNetAP-10    & $47.2_{\pm1.3}$ & $40.6_{\pm0.4}$ & $49.1_{\pm0.7}$ & $49.3_{\pm0.2}$ & $47.1_{\pm1.1}$ & $48.3_{\pm1.0}$ & -        & ${56.3_{\pm1.0}}^*$ & -          & $\boldsymbol{70.2_{\pm0.8}}$  \\
                                                       & ResNet-18        & $47.9_{\pm1.8}$ & $39.6_{\pm1.0}$ & $48.3_{\pm1.2}$ & $50.1_{\pm0.5}$ & $46.2_{\pm0.6}$ & $48.3_{\pm0.8}$ & -        & $57.1_{\pm0.6}$ & ${68.5 _{\pm 0.7}}^*$ & $\boldsymbol{71.3_{\pm0.2}}$ \\ \hline
$100$                                                    & ConvNet-6      & $52.2_{\pm0.4}$ & $45.1_{\pm0.5}$ & $54.4_{\pm1.1}$ & $53.4_{\pm0.3}$ & $55.0_{\pm1.3}$ & $53.2_{\pm0.9}$ & -        & ${61.1_{\pm0.7}}^*$ & -          & $\boldsymbol{67.2_{\pm0.2}}$ \\
$(7.7\%)$                                                & ResNetAP-10    & $59.4_{\pm1.0}$ & $54.8_{\pm0.2}$ & $61.7_{\pm0.9}$ & $58.3_{\pm0.8}$ & $56.4_{\pm0.8}$ & $56.1_{\pm0.9}$ & -        & ${64.5_{\pm0.2}}^*$ & -          & $\boldsymbol{76.7_{\pm0.3}}$ \\
                                                       & ResNet-18        & $61.5_{\pm1.3}$ & $50.4_{\pm0.4}$ & $59.3_{\pm0.7}$ & $58.9_{\pm1.3}$ & $60.2_{\pm1.0}$ & $58.3_{\pm1.2}$ & -        & ${65.7_{\pm0.4}}^*$ & -          & $\boldsymbol{77.3_{\pm0.7}}$ \\ \hline
\end{tabular}}
% \vspace{-0.3cm}
\end{table}

\subsection{Information-Guided Diffusion Sampling} \label{sec:diff}
Based on the discussion above, to maximize $\mathrm{I}(X; \hat{X})$ for an encoder $f_{\boldsymbol{\theta}}(\cdot)$, one can train it to maximize the following objective function: 
\begin{align}
    \mathcal{J}_{\mathrm{VE}} =  - &\mathbb{E}_{\hat{X},Y} \log\left[\frac{\exp(\langle \hat{x}_i, \hat{x}_i\rangle / \tau)}{\sum_k \exp(\langle \hat{x}_i, \hat{x}_k\rangle / \tau)}\right]  \; + \; \lambda \; \mathbb{E}_{\hat{X},Y}\mathrm{KL}(\sigma(\hat{X})||Q^Y),
\end{align}
where $\lambda$ is a hyperparameter that balances the effects of two terms in the objective function. We note that VE reduces to MOCO \citep{he2020momentum} when $\lambda =0$.

By maximizing $\mathcal{J}_{\mathrm{VE}}$, we can effectively train $f_{\boldsymbol{\theta}}(\cdot)$. The pseudo-code for this training procedure is provided in \cref{alg:VE}.

Once $f_{\boldsymbol{\theta}}(\cdot)$ is trained, we proceed to the next step. We freeze the parameters of $f_{\boldsymbol{\theta}}(\cdot)$ and train the classifier $g_{\boldsymbol{\psi}}(\cdot)$ using the standard cross-entropy (CE) loss.
Once $f_{\boldsymbol{\theta}}(\cdot)$ and $g_{\boldsymbol{\psi}}(\cdot)$ are trained, they are fixed and used during the sampling process of the diffusion model as discussed in the next subsection.

To recap, our primary objective was to maximize the function $\mathrm{I}(X;Y) + \beta \, \mathrm{H}(X | Y)$ (see \cref{eq:ultimatediff}) during the sampling process of the diffusion model. To achieve this, we derived variational estimates for $\mathrm{I}(X;Y)$ (\cref{I_VIA}) and $\mathrm{H}(X | Y)$ (\cref{H_VIA}) by leveraging the training of a variational estimator (VE). Using these estimates, the objective function in \cref{eq:ultimatediff} is reformulated as:
\begin{align} \label{eq:IGDS}
    \mathcal{L}_{IGDS} =  \mathbb{E} \log P_{Y|\hat{Y}} + \beta \mathrm{I}(\hat{X} ;\sigma(\hat{X})| Y).
\end{align}
The objective function $\mathcal{L}_{IGDS}$ can be maximized during the sampling process of any diffusion model. As an example, \cref{alg:IGDS} illustrates how our method can be integrated with the denoising diffusion probabilistic model (DDPM) \citep{ho2020denoising}. We refer to the resulting DM-based sampling method as information-guided diffusion sampling (IGDS).

\begin{table}[]
\centering
\caption{Comparing model's performance in terms of accuracy on ImageNette validation set. All results are evaluated under the resolution $256\times 256$. We use \textbf{bold} number and asterisk ($^*$) to denote the best and the second best results, respectively.}\label{Tab:NetteResults}

\resizebox{\textwidth}{!}{\begin{tabular}{c|ccccc|ccc}
\hline
Model & \multicolumn{5}{c|}{ResNetAP-10}                         & \multicolumn{3}{c}{ResNet-18}       \\ \cline{2-9} 
% IPC   & Random   & \makecell{DiT \\ \citep{sun2024diversity}}     & \makecell{DM\\ \citep{zhao2023dataset}} & \makecell{MiniMax\\ \citep{gu2024efficient}}  & Ours     & \makecell{RDED\\ \citep{sun2024diversity}}& \makecell{$SRe^2L$ \\ \citep{yin2023squeeze}}& Ours     \\ \hline
IPC   & Random   & \makecell{DiT }     & \makecell{DM} & \makecell{MiniMax}  & Ours     & \makecell{RDED}& \makecell{$SRe^2L$ }& Ours     \\ \hline
1     & $26.7_{\pm1.0}$ & $27.3_{\pm0.9}$ & -        & ${30.5_{\pm0.8}}^*$ & $\boldsymbol{39.6_{\pm1.3}}$ & ${35.8_{\pm1.0}}^*$ & $19.1_{\pm1.1}$ & $\boldsymbol{35.9_{\pm0.7}}$ \\
10    & $54.3_{\pm1.6}$ & $59.1_{\pm0.7}$ & $60.8_{\pm0.6}$ & ${62.0_{\pm0.2}}^*$ & $\boldsymbol{68.3_{\pm0.2}}$ & ${61.4_{\pm0.4}}^*$ & $29.4_{\pm3.0}$ & $\boldsymbol{64.3_{\pm0.6}}$ \\
20    & $63.5_{\pm0.5}$ & $64.8_{\pm1.2}$ & $66.5_{\pm1.1}$ & ${66.8_{\pm0.4}}^*$ & $\boldsymbol{72.4_{\pm0.7}}$ & -        & -        & $\boldsymbol{70.9_{\pm0.3}}$ \\
50    & $76.1_{\pm1.1}$ & $73.3_{\pm0.9}$ & $76.2_{\pm0.4}$ & ${76.6_{\pm0.2}}^*$ & $\boldsymbol{81.0_{\pm0.5}}$ & ${80.4_{\pm0.5}}^*$ & $40.9_{\pm0.3}$ & $\boldsymbol{81.2_{\pm0.4}}$ \\ \hline
\end{tabular}}
% \vspace{-0.3cm}
\end{table}
\section{Experiments} \label{sec:exp}
\subsection{Experimental Setup}

\noindent $\bullet$ \textbf{Implementation Details of IGDS.}
We adopt the pre-trained DDPM model \citep{dhariwal2021diffusion} and use the pre-trained MoCo model \citep{he2020momentum} as the encoder. To smooth gradients, we follow \citep{ma2024elucidating} and replace the ReLU activation function with SoftPlus \citep{dugas2000incorporating} using $\beta = 3$. A linear classifier is then trained on top of the frozen encoder.  During the DM sampling, we set the temperature to $\tau=0.07$ and run the diffusion process for 250 steps in all experiments. Following \citep{sun2024diversity}, we enhance sample information by merging four images from the same class into a single composite image. We report the evaluation protocol in \cref{Sec:EvaluationProtocol}, and also \cref{Sec:IGDS_Samples} provides sample images from the distilled datasets generated by IGDS. All experiments are conducted on a single NVIDIA V100 GPU. Full implementation details, including code and configurations, are available in our \textit{GitHub} repository.

\begin{wraptable}{h}{0.6\textwidth}
    \centering
    \vspace{-0.3cm}
    \caption{Comparing model's performance in terms of accuracy on Tiny ImageNet validation set. All results are evaluated under the resolution $64\times 64$. We use \textbf{bold} number and asterisk ($^*$) to denote the best and the second best results, respectively.}
    % \vskip -0.15in
    \label{Tab:TinyImgNetRes}
    \resizebox{0.6\textwidth}{!}{\begin{tabular}{c|cccc}
    \hline
    Model & \multicolumn{4}{c}{ConvNet-4}                 \\ \cline{2-5} 
    IPC   & Random   & \makecell{IDM  \\ \citep{zhao2023improved}}   & \makecell{RDED \\ \citep{sun2024diversity}}    & Ours     \\ \hline
    1     & $6.7_{\pm0.4}$  & $10.1_{\pm0.2}$ & $\boldsymbol{12.0_{\pm0.1}}$ & ${11.9_{\pm0.3}}^*$ \\
    10    & $17.6_{\pm0.3}$ & $21.9_{\pm0.3}$ & ${39.6_{\pm0.1}}^*$ & $\boldsymbol{40.7_{\pm0.3}}$ \\
    50    & $22.4_{\pm0.2}$ & $27.7_{\pm0.3}$ & ${47.6_{\pm0.2}}^*$ & $\boldsymbol{50.3_{\pm0.2}}$ \\ \hline
    Model & \multicolumn{4}{c}{ResNet-18}                 \\ \cline{2-5} 
    IPC   & Random   & \makecell{$SRe^2L$\\ \citep{yin2023squeeze}}    & \makecell{RDED \\  \citep{sun2024diversity}}   & Ours     \\ \hline
    1     & $2.2_{\pm0.4}$  & $2.6_{\pm0.1}$  & ${9.7_{\pm0.4}}^*$  & $\boldsymbol{9.8_{\pm0.4}}$  \\
    10    & $14.6_{\pm0.2} $& $16.1_{\pm0.2} $& $\boldsymbol{41.9_{\pm0.2}}$ & ${41.2_{\pm0.1}}^*$ \\
    50    & $35.6_{\pm0.3}$ & $41.1_{\pm0.4}$ & ${58.2_{\pm0.1}}^*$ & $\boldsymbol{60.1_{\pm0.5}}$ \\ \hline
    \end{tabular}}
\vspace{-0.2cm}
\end{wraptable}

\noindent $\bullet$ \textbf{Datasets.} To evaluate the effectiveness of the proposed method, we conduct experiments mainly on several benchmark datasets. We select ImageNet-1K \citep{5206848} and three well-known subsets of ImageNet: ImageNette, ImageWoof, and Tiny ImageNet \citep{le2015tiny}. ImageNet is a large-scale visual recognition dataset containing approximately 1.2 million training images and 50,000 validation images. ImageNette, a subset of ImageNet, provides a smaller and more manageable dataset for testing deep learning models, while ImageWoof focuses on 10 dog breeds, offering a fine-grained classification task. Both ImageNette and ImageWoof use a spatial resolution of $224\times224$. Tiny ImageNet, on the other hand, is a small, balanced subset of ImageNet, with its training set consisting of 200 classes---each class containing 500 samples resized to a spatial resolution of $64\times64$. 

\noindent $\bullet$ \textbf{Network Architectures.} Following previous work \citep{cazenavette2022dataset, cui2023scaling, guo2024datm}, we use ConvNet-4 \citep{lecun1998gradient} for the Tiny ImageNet dataset and ConvNet-6 for the ImageWoof dataset. Additionally, we employ ResNetAP-10, a variant of ResNet-10 where all pooling layers are replaced with average pooling, and ResNet-18 for all experiments. 

\subsection{How to select $\beta$ in \cref{eq:IGDS}} \label{sec:beta}

\begin{figure}
    \begin{center}
    \vspace{-1.0cm}
    \includegraphics[width=0.8\textwidth]{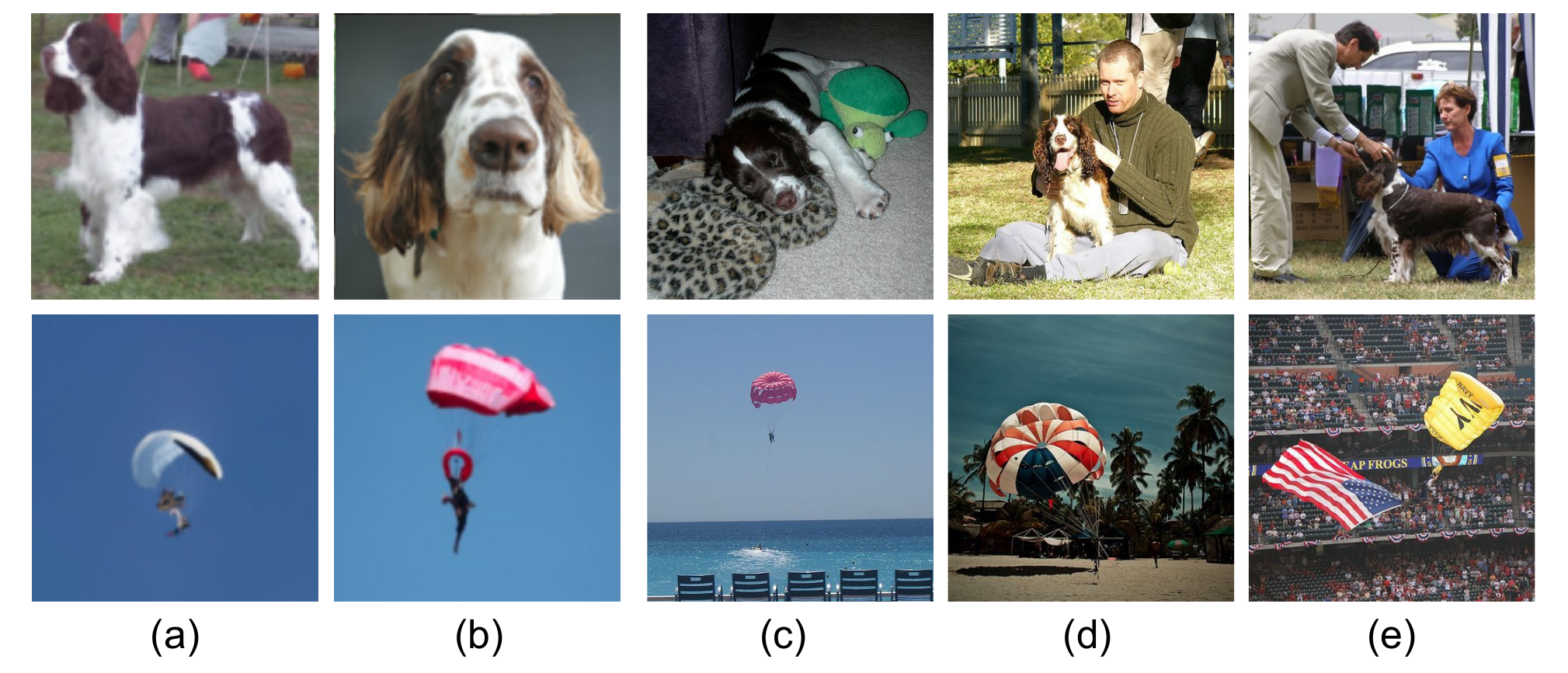}
    \end{center}
     \caption{Illustration of prototype and contextual information. ($i$) Top: English Springer; ($ii$) Bottom: Parachute. The first two columns show synthetic images with low contextual information, while the last three columns display natural images from the same classes.}
    % \vspace{-1.5cm}
    \label{fig:CentriodsVis}
\end{figure}

As discussed in \cref{sec:method}, the parameter $\beta$ in \cref{eq:IGDS} should be selected based on the IPC. To illustrate this, we generate multiple subsets of Tiny ImageNet with varying $\underline{\mathrm{H}}(X|Y)$ and IPC values. To control $\underline{\mathrm{H}}(X|Y)$, we apply a weighted sampling method, which is detailed in the Supplementary Materials. We then train ConvNet-4 on these subsets and report the classification accuracies. Finally, we plot the relationship between $\underline{\mathrm{H}}(X|Y)$ and model validation accuracy across different IPC settings in \cref{fig:onecol}. As observed, higher IPC settings require greater contextual detail, while lower IPC settings benefit from a stronger emphasis on prototype information.

\subsection{Comparison with State-of-the-art Methods}
We first report the experimental results for ImageWoof in \cref{Tab:ImageWoofResults}. As seen, at a low IPC setting (IPC-1), the performance of all generative-based dataset distillation methods, except IDGS, is close to that of a randomly selected subset. However, as the IPC increases, the performance gap between the baseline methods and the random selection also increases. Nevertheless, none of the baseline methods outperforms IDGS.

We also present the experimental results for ImageNette and Tiny ImageNet in \cref{Tab:NetteResults} and \cref{Tab:TinyImgNetRes}, respectively. The results follow a similar trend to those on the ImageWoof dataset. 

We defer the ImageNet-1K, Cifar-10 and cross-architecture results to the appendix.
\subsection{Semantic Meaning of Prototype and Contextual Information} 
\label{sec:semantic}

\begin{figure}
    \centering
    % \vspace{-0.5cm}
    \includegraphics[width=0.55\linewidth]{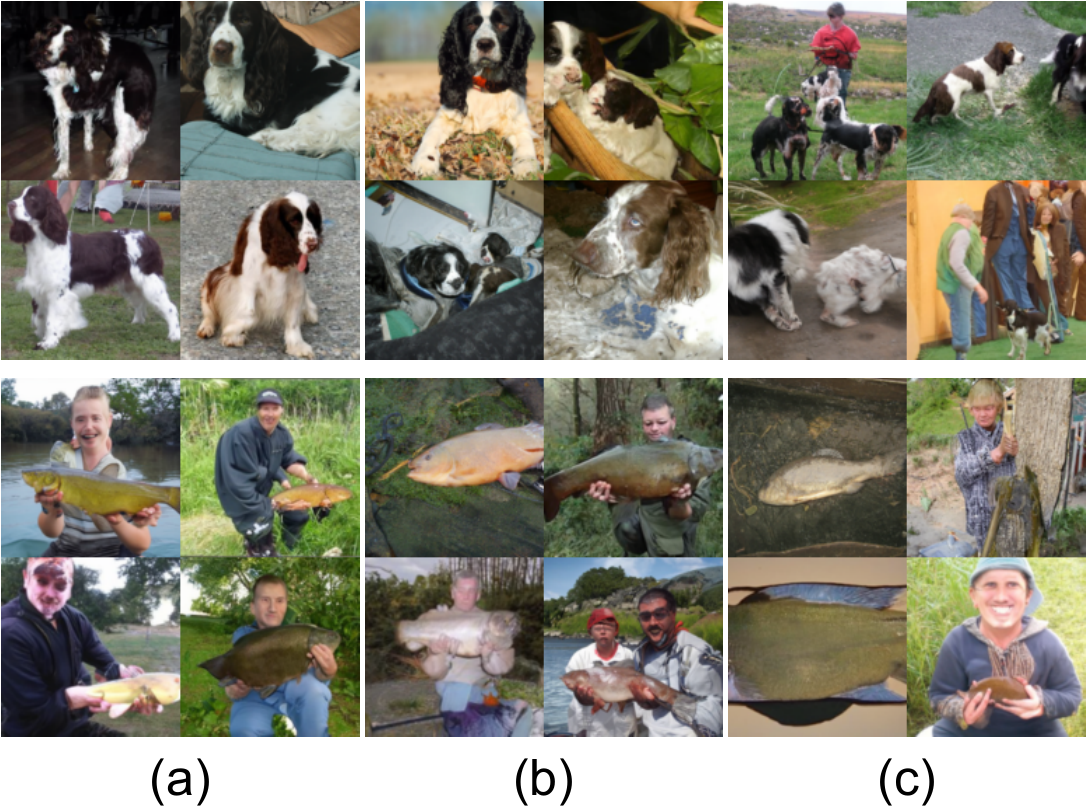}
    \caption{Generated images of two classes: English Springer (first row) and Tench (second row), with varying $\beta$ values. Columns (a), (b), and (c) correspond to  $\beta=\{0, 0.1,0.5\}$, respectively.}
     \label{fig:DifferentBeta}
\end{figure}

To better understand the semantic meaning of prototype and contextual information, we visualize samples generated by IGDS with $\beta = 0$, where \textit{only} prototype information is maximized during the DM sampling process. These synthetic images are shown in the first two columns of \cref{fig:CentriodsVis} (columns (a) and (b)).  For direct contrast, three natural images randomly selected from the same class are displayed in the last three columns (columns (c) to (e)). As observed, the synthetic images with low contextual information feature plain backgrounds and minimal context, whereas the natural images exhibit richer contextual details.

\section{Cross-architecture performance}
\begin{table}[h!]
\centering
\caption{Cross-architecture performance}
\label{Tab:Cross}
\resizebox{0.8\linewidth}{!}{\begin{tabular}{c|cccc}
\hline
Method & ResNet-101 & MobileNet-V2 & EfficientNet-B0 & SwinT \\ \hline
Minimax-IGD  & 53.4      & 39.7         & \textbf{48.5}            & 44.8  \\
Minimax-ours & \textbf{53.6}      & \textbf{39.9}         & 48.3            & \textbf{45.9}  \\ \hline
\end{tabular}}
\end{table}

To evaluate the robustness of our distilled datasets across diverse model families, we follow the IGD protocol and measure performance when training four different architectures on the same distilled coresets. \cref{Tab:Cross} summarizes the test accuracies achieved by Minimax-IGD and our method under the IPC‑10 setting.

\section{CIFAR-10 Results}
\label{Sec:Cifar10}
We evaluated the performance of our approach and baseline methods on the CIFAR-10 dataset using a ResNet-18, as shown in the Tab. \ref{Tab:Cifar}. As observed, our method achieves comparable or superior performance to previous state-of-the-art methods on this low-resolution dataset.
\begin{table}[h!]
% \vspace{-0.4cm}
\centering
\caption{Performance comparison over ResNet-18 on CIFAR-10}
\label{Tab:Cifar}
% \vspace{-0.35cm}
\resizebox{0.6\linewidth}{!}{\begin{tabular}{cc|ccc|c}
\hline
IPC&Test Model& $SRe^2L$ & RDED & DIT-IGD & Ours  \\ \hline
10 &ResNet-18 & 29.3 & \textbf{37.1} & 35.8    & 37.0 \\
50 &ResNet-18 & 45.0 & 62.1 & 63.5    & \textbf{64.9} \\ \hline
\end{tabular}}
\end{table}

\subsection{Ablation Study on $\beta$ in IGDS}
\label{Sec:Beta_IGDS}

In this section, we study the impact of $\beta$ on IGDS, the performance of the distilled dataset under different IPC settings, and its effect on the generated images. To this end, we first examine how $\beta$ influences the semantic meaning of generated images, as shown in \cref{fig:DifferentBeta}. Specifically, we generate 24 synthetic images for the classes English Springer and Tench, displayed in the first and second rows of \cref{fig:DifferentBeta}, respectively. Columns (a), (b), and (c) correspond to $\beta = \{0, 0.1, 0.5\}$. When $\beta = 0$, the generated images contain minimal contextual information. For example, the English Springer images primarily depict the dog itself, while the Tench images consistently depict a person holding the fish. As $\beta$ increases, more contextual elements are incorporated into the generated images, leading to a greater diversity in semantic meaning. This effect is particularly noticeable in the English Springer images, where the background becomes richer compared to those generated with $\beta = 0$. A similar trend can be observed in the Tench images, where additional contextual details emerge as $\beta$ increases. More images generated by IGDS with different $\beta$ values are presented in \cref{Sec:IGDS_Samples}.

\begin{wrapfigure}{h}{0.6\textwidth}
    \begin{center}
    \vspace{-1.2cm}
    \includegraphics[width=0.6\textwidth]{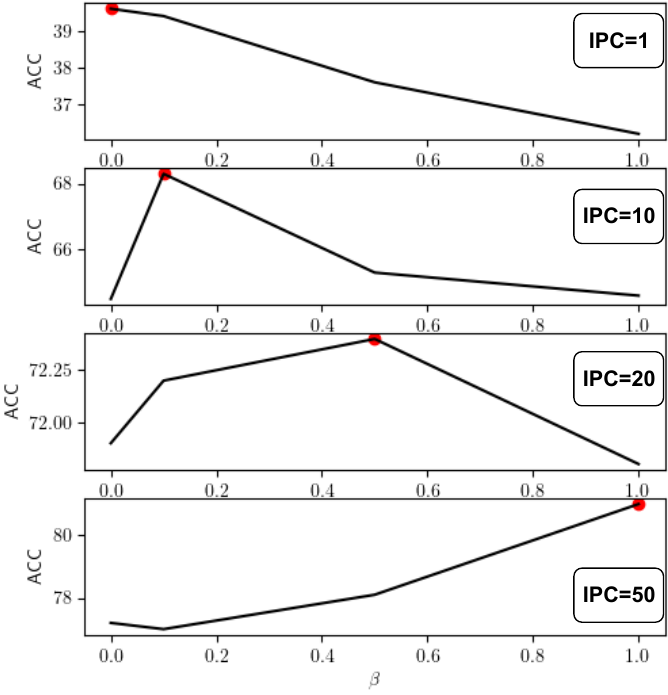}
    \end{center}
    \caption{The model's accuracy on the distilled dataset Vs. $\beta$. As observed, lower IPC settings favor smaller $\beta$ values, whereas higher IPC settings require an increased $\beta$ value accordingly.}
    \label{fig:DifferentIPCfigure}
    \vspace{-1.5cm}
\end{wrapfigure}

The optimal value of $\beta$ should be empirically determined for different IPC settings. To illustrate this, \cref{fig:DifferentIPCfigure} shows the test accuracy of the model as a function of $\beta$ under varying IPC values. As observed, higher IPC settings benefit from a larger $\beta$, aligning with the findings discussed in \cref{sec:beta}.

\section{Conclusions and Future Works}
In this work, we addressed the limitations of diffusion model-based dataset distillation in low-IPC settings through an information-theoretic approach. We identified prototype information $\mathrm{I}(X;Y)$ and contextual information $\mathrm{H}(X | Y)$ as essential components and proposed maximizing $\mathrm{I}(X;Y) + \beta \mathrm{H}(X | Y)$ during sampling, with $\beta$ adapted to IPC. To handle intractability, we introduced variational estimations using a deep neural network. Our proposed method, information-guided diffusion sampling (IGDS), seamlessly integrated with diffusion models and achieved state-of-the-art performance on Tiny ImageNet and ImageNet subsets, particularly in low-IPC regimes.

Despite the theoretical contributions and promising results of the proposed information-guided diffusion sampling (IGDS) method, this work has several limitations. First, like previous studies, we use a pretrained diffusion model as the prior distribution for natural images. While this approach is intuitive, its optimality for dataset distillation remains unverified. In addition, it restricts applicability by requiring a pretrained diffusion model for the target dataset. Second, during the IGDS process, gradients must be backpropagated through both the classifier and encoder to guide the diffusion process, increasing computational costs. We report the runtime analysis in \cref{Sec:RunningTime}. Addressing these limitations is a direction for future work.
\clearpage

\bibliography{main}
\bibliographystyle{tmlr}

\clearpage
\appendix
\section*{Appendix}
\section{Weighed Sampling to Generated Subsets with Different \texorpdfstring{$\underline{\mathrm{H}}(X|Y)$}{} values}
\label{Appendix:WeighedSampling}
In this section, we describe how to perform weighted sampling to generate the subset of a dataset with different $\underline{\mathrm{H}}(X|Y)$ values. Given a dataset $\mathcal{D}$ of size $n$ with $C$ classes, $\mathcal{D} = \{(\boldsymbol{x}_i, y_i)\}_{i=1}^n$, where each $\boldsymbol{x}_i \in \mathbb{R}^d$ and $y_i \in [C]$, a pretrained encoder $f(\cdot)$ trained using the VE method on $\mathcal{D}$, and a classifier  $g(\cdot)$, we first filter out all misclassified samples to ensure that the remaining samples’ contextual information can be captured by  $g(f(\cdot))$. In the Shannon sense, the contextual information for each sample within class $y$ is quantified using the $\mathrm{KL}$ divergence $\mathrm{KL}(\sigma(x)||Q^y)$, where $\sigma(x)$ is a probability vector obtained by applying the softmax function to the feature map $\hat{x}$, and $Q^y$ is estimated as $\frac{1}{|D_y|}\sum_{x\in D_y} H$. Given a target $\alpha_{\underline{H}(\cdot|y)}$ value, we compute the probability $\tilde{P}_{X|Y}(\cdot| y)$ of each sample being selected as follows:

\begin{align}
    \tilde{P}_{X|Y}(x| y) = \frac{\exp({-(\mathrm{KL}(\sigma(x)||Q^y)-\alpha_{\underline{H}(\cdot|y)})^2})}{\sum_{x' \in D^y} \exp({-(\mathrm{KL}(\sigma(x')||Q^y)-\alpha_{\underline{H}(\cdot|y)})^2})},
\end{align}
then, IPC samples are drawn from each class according to the probability $\tilde{P}_{X|Y}(\cdot|y)$. We visualize the samples that map to $Q^Y$ using the pretrained encoder $f(\cdot)$ in Figure \ref{fig:CentriodsVis} to enhance understanding of the semantic meaning of prototype information and contextual information.

\section{Proof of Propositions} \label{sec:proofs}
\subsection{Proof of \cref{Lemma:ProofInjectiveEq}}
\label{Appendix:ProofInjective}

\begin{proof}
    We first prove that for any injective function $f$, 
    \begin{align}
        \mathrm{I}(X; Y) = \mathrm{I}(f(X); Y),
    \end{align}
    To do so, we begin by expanding the mutual information $\mathrm{I}(\cdot ; \cdot)$, and introducing the variable $Z = f(X)$:
    \begin{align}
    \mathrm{I}(X; Y) -\mathrm{I}(f(X); Y) &=\mathrm{H}(Y|f(X))- \mathrm{H}(Y|X)  \\
    &= \mathrm{H}(X|f(X)).
    \end{align}
    Since $f$ is injective, for any output $z = f(X)$, there exists a unique $x$ such that $f(x) = z$. Therefore, 
    \begin{align}
      P(X=x|f(X)=z) = \begin{cases}
      1 & \text{if $z = f(x)$},\\
      0 & \text{otherwise}.
    \end{cases} 
    \end{align}
    The conditional entropy is then given by:
    \begin{align}
        \mathrm{H}(X|f(X)) & = \mathbb{E}_{f(X)} [\mathrm{H}(X|f(X) = z)]\\
        & = \mathbb{E}_{f(X)} 0\\
        & = 0.
    \end{align}
    Thus if $f$ is injective, we conclude that $I(Y,f(X)) = I(Y,X)$.
    
    Next, we show that for a matrix $\theta\in \mathbb{R}^{m,n}, m\geq n$, if $\theta$ has full column rank, then the linear mapping $\{\hat{X}\rightarrow \hat{Y}$, where $\hat{Y} = \theta \hat{X}\}$, is injective.

    A function is injective if:
    \begin{align}
        \theta x_1 = \theta x_2 \Rightarrow x_1 = x_2,
    \end{align}
    Rearranging, introducing $v = x_1-x_2$ we get: $\theta v =0$.
    For injectivity, we must show that the only solution to $\theta v=0$ is $v=0$.
    
    The set of all solutions to $\theta v =0$ is the null space of $\theta$, denoted as:
    \begin{align}
        Null(\theta) = \{v\in \mathbb{R}^n|\theta v = 0\}.
    \end{align}
    If the linear mapping is injective, the only vector in the null space must be the zero vector, \ie $Null(\theta) = \{0\}$, which means $\theta$ has full column rank.
\end{proof}

We verify that the classifier's matrix after training has full column rank.

\subsection{Proof of \cref{prop:I}}

$\min_\theta \mathrm{I}(Y;X|\hat{X}) \equiv \max_\theta \mathrm{I}(X;\hat{X})  $
\begin{proof}
    \begin{align}
        \mathrm{I}(Y;X|\hat{X})=&\mathrm{I}(Y;X|\hat{X}) \\
        = &\mathrm{H}(X|\hat{X}) - \mathrm{H}(X|Y,\hat{X})\\
         =& \mathrm{H}(X)- \mathrm{H}(X|Y) + \nonumber\\
        &\mathrm{H}(X|\hat{X})- \mathrm{H}(X) \\
         =& \mathrm{I}(X;Y) - \mathrm{I}(X;\hat{X}),
    \end{align}
where $\mathrm{I}(X;Y)$ is a constant, which only depends on the nature of the sampling process, \ie how the dataset is collected and constructed.
\end{proof}

\subsection{Proof of \cref{lem:soft}}
Assume that the feature representation $\hat{X}$ has zero mean. Then, $\mathrm{I}(X;\hat{X}|Y) = \mathrm{I}(X;\sigma(\hat{X})|Y)$. 

\begin{proof}
    The softmax function for an input vector $x\in \mathbb{R}^N$ is define as

\begin{align}
    softmax(x)[i] = \frac{e^{x[i]}}{\sum_{j\in [N]}e^{x[j]}}.
\end{align}

Following the proof of \cref{Lemma:ProofInjectiveEq} in \cref{Appendix:ProofInjective}, we aim to show that the softmax function is injective if its domain is in the subspace with zero mean. Assume two vectors $\ x;y \in \mathbb{R}^N$, such that 
\begin{align}
    \sum_{i\in[N]} x_i = 0,\quad \sum_{i\in[N]}y_i = 0,\\
    \frac{e^{x[i]}}{\sum_{j\in [N]}e^{x[j]}} = \frac{e^{y[i]}}{\sum_{j\in [N]}e^{y[j]}}. \label{Eq:softmaxEq}
\end{align}

Rewriting \cref{Eq:softmaxEq}, 

\begin{align}
    e^{x[i]}{\sum_{j\in [N]}e^{y[j]}} = e^{y[i]} {\sum_{j\in [N]}e^{x[j]}},\ \  \forall i\in[N]. \label{Eq:SoftMaxRearrange}
\end{align}

We define the ratio of $\sum_{j\in [N]}e^{x[j]}$ and $\sum_{j\in [N]}e^{y[j]}$ as:

\begin{align}
    \Theta = \frac{\sum_{j\in [N]}e^{x[j]}}{\sum_{j\in [N]}e^{y[j]}}. \label{Eq:SoftMaxRatio}
\end{align}

Substitute the \cref{Eq:SoftMaxRatio} into \cref{Eq:SoftMaxRearrange} and take logarithm on both sides:

\begin{align}
    x_i = \log \Theta + y_i.
\end{align}
Since both $x$ and $y$ have zero mean, we have:

\begin{align}
    \sum_{i\in[N]} x_i =\sum_{i\in[N]} \log \Theta + y_i = 0,\\
    N\log \Theta = 0,\  \Theta = 1.
\end{align}

Thus $x_i = y_i,\ \forall i\in[N]$. 
\end{proof}

\subsection{Proof of \cref{prop:H}}
\label{Appendix:ProofofEquOptProb}
For a encoder $f$ parametrized by $\theta$.\\

$\min_\theta \mathrm{H}(X|\hat{X},Y) \equiv \max_\theta \mathrm{I}(X;\hat{X})$
\begin{proof}
    \begin{align}
        \mathrm{I}(X;\hat{X}) &= \mathrm{H}(X) - \mathrm{H}(X|\hat{X})\\
        &= \mathrm{H}(X) - \mathrm{H}(X|\hat{X},Y) \label{eq:addCondition}
        % & = \mathrm{H}(X) -  \mathrm{H}(X|\hat{X},Y),
    \end{align}
where $\mathrm{H}(X)$ is a constant, which equals the amount of information in the dataset, \cref{eq:addCondition} is due to $Y\rightarrow X \rightarrow \hat{X}$ forms a Markov chain. 
\end{proof}

\subsection{Proof of \cref{Lemma:MaxHX}}
\label{Sec:ProofMaxHX}
\begin{proof}
    Consider a random variable $Y$, with $N$ classes. Its entropy is given by 
    \begin{align}
        \mathrm{H}(Y) = -\sum_{n\in [N]}P_n \log P_n.
    \end{align}
    Without losing generality, suppose we split the first sample point $y_1$ into two sample points $\hat{y}_a$ and $\hat{y}_b$, such that 
    \begin{align}
        P[{y_1}] &= P[{\hat{y}_a}]+P[{\hat{y}_b}], \\
        &s.t.\ P[{\hat{y}_a}]>0; P[{\hat{y}_b}]>0,
    \end{align}
    This transformation produces a new random variable $\hat{Y}$ with $N+1$ sample points. The change in entropy is then
    \begin{align}
        &\mathrm{H}(\hat{Y})-\mathrm{H}(Y) \\
        = &-P[\hat{y}_b]\log P[\hat{y}_b]-P[\hat{y}_a]\log P[\hat{y}_a]+ P[y_1]\log P[y_1] \\
        = & -P[\hat{y}_b]\log P[\hat{y}_b]-P[\hat{y}_a]\log P[\hat{y}_a]+ \{P[\hat{y}_b]+P[\hat{y}_a]\}\log \{ P[\hat{y}_b]+P[\hat{y}_a] \}\\
        = &-P[\hat{y}_b]\big\{\log P[\hat{y}_b]-\log \{ P[\hat{y}_b]+P[\hat{y}_a] \}\big\}-P[\hat{y}_a]\big\{\log P[\hat{y}_a]-\log \{ P[\hat{y}_b]+P[\hat{y}_a] \}\big\}\\
         >  &0.
    \end{align}
    Thus $\mathrm{H}(\hat{Y}) > \mathrm{H}(Y)$, that is, splitting a single sample point into two distinct points increases the entropy of a random variable. In other words, given a set of samples, the entropy $\mathrm{H}(Y)$ is maximized when each sample is assigned to a unique class. Conversely, for a fixed number of sample points, the entropy is maximized when their probabilities are uniformly distributed. Therefore, the entropy $\mathrm{H}(Y)$ is maximized if the problem is formulated as instance discrimination.
\end{proof}

\section{Further discussion on VE and information bottleneck}
\label{Appendix:VEandBI}
In this section, we discuss the relationship between the maximized mutual information method and information bottleneck, and provide an alternative approach to derive the objective function for the VE method.

\subsection{Relationship Between VE and Information Bottleneck}
\begin{figure}[h!]
\begin{center}
\centerline{\includegraphics[width=\columnwidth]{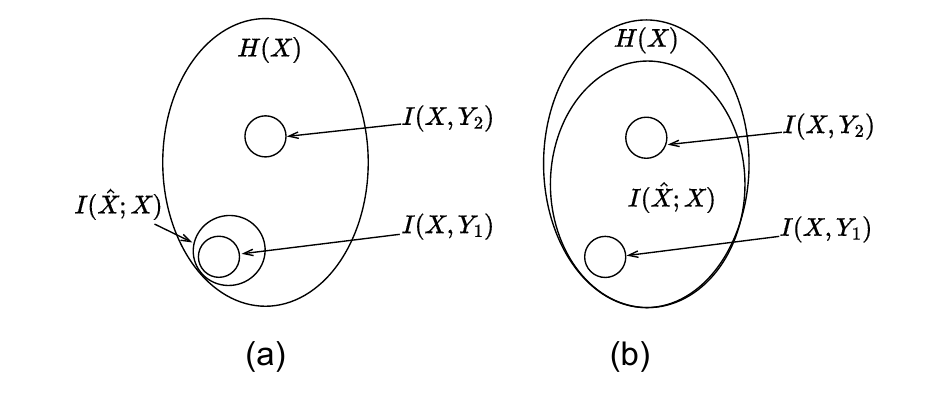}}
\caption{Mutual information between $\hat{X}$ and $X$ for models trained with objectives of (a) information bottleneck and (b)  VE (ours).}
\label{Fig:InfoRelation}
\end{center}
\end{figure}
We depict the Venn diagram of which show the relationships between the $\mathrm{H}(X)$, prototype information $\mathrm{I}(X;Y)$, contextual information $\mathrm{H}(X|Y)$ and $\mathrm{I}(\hat{X}; X)$ by the encoder trained by information bottleneck (a) and VE (b) in \cref{Fig:InfoRelation}. 

Information bottleneck aims to minimize the following objective function:
\begin{align}
    \min \mathrm{I}(X; \hat{X}) - \beta \mathrm{I}(\hat{X}; Y),
\end{align}
which can be interoperated as finding a compressed representation $\hat{X}$ of $X$ that retains as much information about $Y$ as possible, while minimizing the information retained from $X$.

While the target of VE differs from the information bottleneck, as it aims to maximize the mutual information $ \mathrm{I}(X; \hat{X})$, as though, the compressed representation $\hat{X}$ retains as much information about $X$ as possible.

\subsection{Alternative Approach to Simplify the Prototype Information}

With a slight abuse of notation, in this section, we refer to $\hat{Y}$  as the label predicted from the feature $\hat{X}$.

\begin{align}
    \mathrm{I}(\hat{X}; Y) &= \mathrm{H}(Y) - \mathrm{H}(Y|\hat{X}) \\
    & \geq \mathrm{H}(Y) -  \mathrm{H}(E) - P(E)\log(|Y|-1) \label{Line:Fano}\\
    & \geq \mathrm{H}(Y) -  \mathrm{H}(E) -  \log(|Y|-1) \mathbb{E}_{X}\Bigg[1-  \sum_{i=1}^{|Y|} P_{Y|X}(i|x)P_{\hat{Y}|X}(i|x)\Bigg] \\
    & =  \mathrm{H}(Y) -  \mathrm{H}(E) - \log(|Y|-1) \mathbb{E}_{X}\Bigg[\sum_{i=1}^{|Y|}  P_{Y|X}(i|x)  \Big[1-P_{\hat{Y}|X}(i|x)\Big]\Bigg] \\
    & \geq \mathrm{H}(Y) -  \mathrm{H}(E) - \log(|Y|-1)  \mathbb{E}_{X}\Bigg[\sum_{i=1}^{|Y|}- P_{Y|X}(i|x)  \log P_{\hat{Y}|X}(i|x) \Bigg] \\
    & = \mathrm{H}(Y) -  \mathrm{H}(E) - \log(|Y|-1)   \mathbb{E}_{X} [\mathrm{H}( P_{Y|X}(i|x), P_{\hat{Y}|X}(i|x))],
\end{align}
where \cref{Line:Fano} follows from \citep{fano1961transmission}, $\log(|Y|-1)$ is a constant, and $\mathrm{H}(E)$ approaches zero when no prediction error occurs.

\section{Running Time}
\label{Sec:RunningTime}

In this section, we report the running time of the IGDS algorithm and compare its efficiency with the MiniMax \citep{gu2024minimax}. To this end, we sampled 100 images with resolution $256\times 256$ for ImageWoof and ImageNette datasets using both methods on the clusters we used; all comparisons are done with one single NVIDIA V100 GPU. The running time is reported in the \cref{Tab:RunningTime}.

\begin{table}[h!]
\centering
\caption{The running time of MiniMax Diffusion and IGDS.}\label{Tab:RunningTime}
\begin{tabular}{ccc}
\hline
Dataset & ImageNette & ImageWoof \\ \hline
MiniMax & 44 mins    & 46 mins    \\
IGDS    & 57 mins    & 53 mins   \\ \hline
\end{tabular}
\end{table}
Compared to MiniMax diffusion, IGDS slightly increases the time complexity, primarily due to the additional VE model required in the distillation process.

\section{Combining with Priors Beyond DDPM}

An important advantage of our approach is its flexibility to integrate with a wide range of generative priors beyond DDPM. To illustrate this capability, we conducted experiments combining our method with the Minimax-DIT prior, which is recognized as a more advanced prior often leading to stronger performance. As shown in \cref{fig:IGD}, when using the Minimax-DIT prior, our method achieves an accuracy of 48.6\% on the ImageWoof dataset under IPC-10. This result demonstrates that our approach not only remains effective when paired with stronger priors but can also be seamlessly combined with alternative generative models to further enhance performance.

\begin{table}[h!]
\centering
\caption{IGD vs. ours when using Minimax-DIT as prior.}
\label{fig:IGD}
\resizebox{0.4\columnwidth}{!}{%
\begin{tabular}{c|cccc}
\hline
IPC          & 1    & 10   & 50   & 100  \\ \hline
Minimax-IGD  & -    & 47.2 & 65.0 & 71.5 \\
Minimax-Ours & 37.6 & \textbf{48.6} & \textbf{65.6} & \textbf{75.3} \\ \hline
\end{tabular}
}
\end{table}

\section{Evaluation Protocol}
\label{Sec:EvaluationProtocol}
We report the evaluation protocol in this section. Three commonly used network architectures are used for evaluation:

\noindent $\bullet$ \textbf{ConvNet-6}, a 6-layer convolutional network, is an extension of ConvNet-3, which is commonly used in previous dataset distillation (DD) works for small-resolution images. To accommodate full-sized 256×256 ImageNet data, we add three additional layers. Each layer contains 128 feature channels, and instance normalization is applied.

\noindent $\bullet$ \textbf{ResNetAP-10} is a 10-layer ResNet variant in which the standard strided convolution is replaced with average pooling for downsampling, allowing for smoother feature aggregation.

\noindent $\bullet$ \textbf{ResNet-18} is an 18-layer ResNet modified to use instance normalization (IN) instead of batch normalization. Since IN performs better than batch normalization under our experimental protocol, we adopt it consistently across all ResNet-18 models.

During the evaluation training, we closely follow the protocols established in \citep{kim2022dataset, gu2024minimax}. Specifically, we use the Adam optimizer with a fixed learning rate of 0.01 across all experiments to ensure consistency in optimization. The number of training epochs for different IPC settings is detailed in Table \ref{Tab:Epochs}. The applied data augmentations are random resize-crop, random horizontal flip, and CutMix \citep{9008296}. 

\begin{table}[h!]
\centering
\caption{Evaluation training epochs across different IPC settings.}\label{Tab:Epochs}
\begin{tabular}{c|ccccc}
\hline
IPC    & 1    & 10   & 20   & 50   & 100  \\ \hline
Epochs & 2000 & 2000 & 1500 & 1500 & 1000 \\ \hline
\end{tabular}
\end{table}

\section{Samples Generated by IGDS}
\label{Sec:IGDS_Samples}
\begin{figure*} [h!]
    \centering
\includegraphics[width=\textwidth]{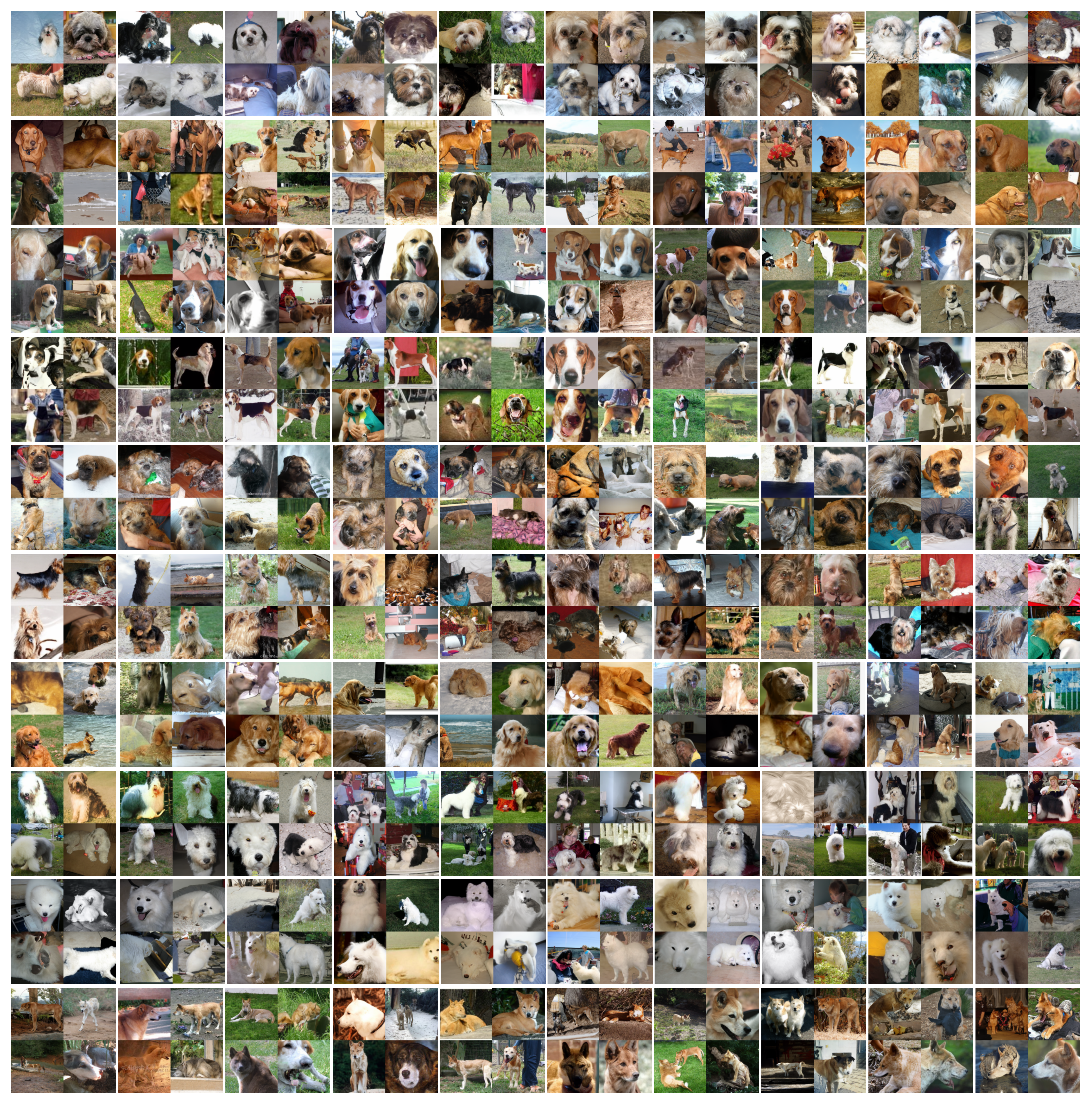}
    \caption{Distilled Image Visualization: ImageNette with IPC 10.}
    \label{fig:ImageNetteSample}
\end{figure*}

\begin{figure*} [h!]
    \centering

\includegraphics[width=\textwidth]{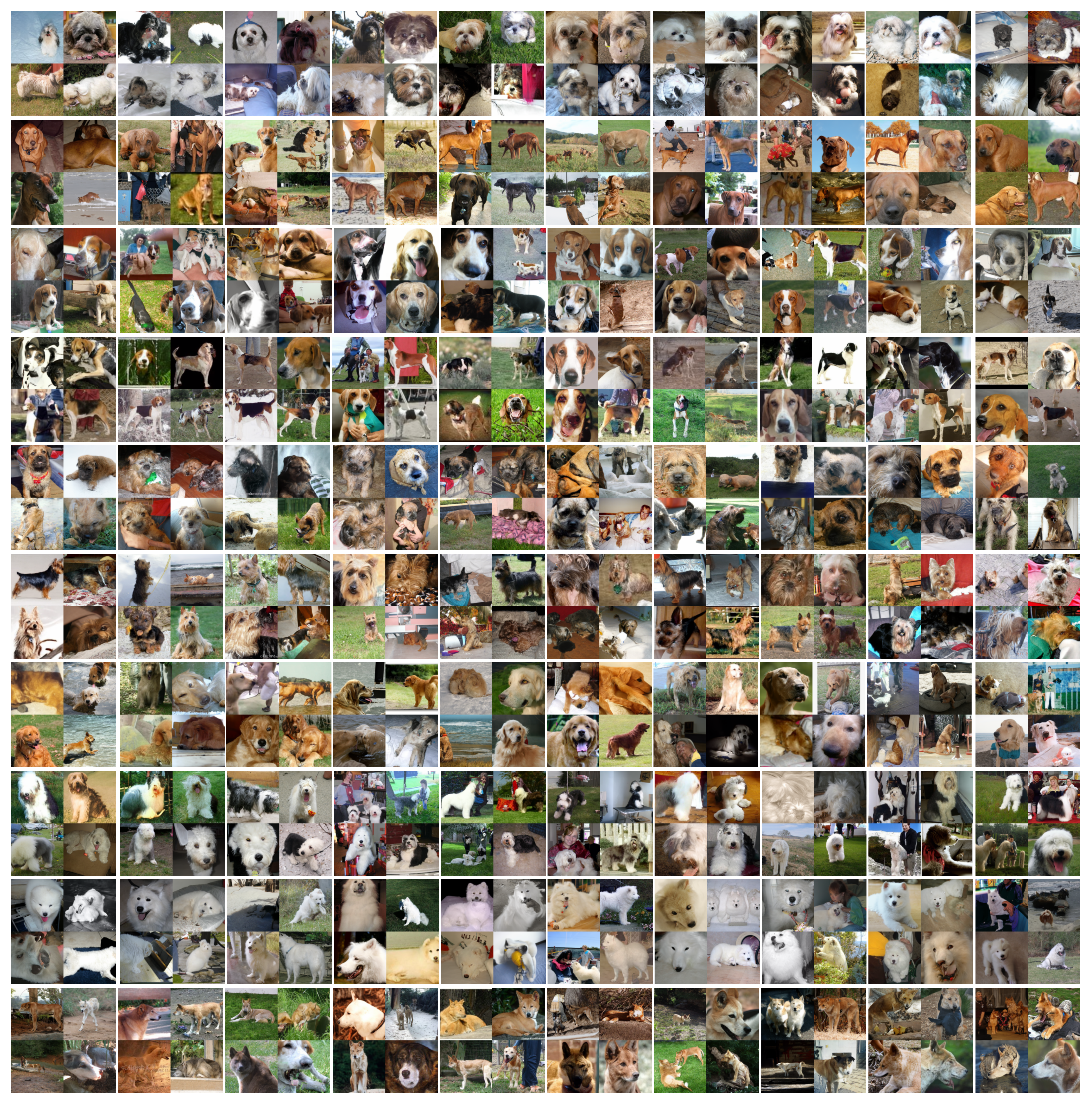}
    \caption{Distilled Image Visualization: ImageWoof with IPC 10.}
    \label{fig:ImageWoofSample}
\end{figure*}

In this section, we provide additional examples of distilled datasets for ImageNette and ImageWoof with IPC 10, shown in \cref{fig:ImageNetteSample} and \cref{fig:ImageWoofSample}, respectively.

\end{document}